\documentclass[conference]{IEEEtran}
\usepackage{times}

\usepackage[numbers]{natbib}
\usepackage{multicol}
\usepackage{subcaption}
\usepackage[bookmarks=true]{hyperref}
\usepackage{hypcap} % fix the links
\usepackage{color,soul}
\usepackage{amsmath}
\usepackage{amsfonts}
\usepackage{graphicx}
\usepackage[table]{xcolor}% http://ctan.org/pkg/xcolor
\usepackage{multirow}
\usepackage{verbatim}
\usepackage[font={small,it}]{caption}
\usepackage{tikz}

\renewcommand{\baselinestretch}{0.965}

\newcommand\blfootnote[1]{%
  \begingroup
  \renewcommand\thefootnote{}\footnote{#1}%
  \addtocounter{footnote}{-1}%
  \endgroup
}

\begin{document}

% paper title
\title{Reasoning About Liquids via \\\vspace{-0.2cm}Closed-Loop Simulation\vspace{-0.5cm}}

% You will get a Paper-ID when submitting a pdf file to the conference system
\author{Connor Schenck and Dieter Fox\vspace{-1.5cm}}

% explicitly state assumptions
% Main points
%several people have started workin on handling liquids
%most use weak models, others showed strong physics based models but only sim not for real (kunze)
%physics based sims for rigid bodies very powerful and models great for reasoning and control contact modeling etc. emo, tetrik, trinkle
%require closed loop to match against real data (also cite tanner's manipulation paper, and peter abiel tracking cloth with physics sim)
%Has not happened for liquid yet, that's what we do

\maketitle

\begin{abstract}
\blfootnote{{\bf Acknowledgement} This work was funded in part by the National Science Foundation under contract numbers NSF-NRI-1525251 and NSF-NRI-1637479.}Simulators are powerful tools for reasoning about a robot's interactions with its environment.
However, when simulations diverge from reality, that reasoning becomes less useful.
In this paper, we show how to close the loop between liquid simulation and real-time perception. We use observations of liquids to correct errors when tracking the liquid's state in a simulator.
Our results show that closed-loop simulation is an effective way to prevent large divergence between the simulated and real liquid states. 
As a direct consequence of this, our method can enable reasoning about liquids that would otherwise be infeasible due to large divergences, such as reasoning about occluded liquid.
\end{abstract}

\IEEEpeerreviewmaketitle

\section{Introduction}

Liquids are ubiquitous in human environments, appearing in many common household tasks.
Recent work in robotics has begun to investigate ways in which robots can reason about and manipulate liquids.
While some research teams have successfully solved liquid pouring tasks using relatively weak models of the physics underlying liquid flow~\cite{yamaguchi2016,schenckc2016c,rozo2013}, other work has shown that physics-based models have the potential to enable far richer understanding of actions involving liquids \cite{kunze2015}. 

Physics-based models are very general tools for enabling robots to reason about their environments.
Work on rigid-body actions using physics-based models has enabled robots to perform a wide variety of tasks \cite{tassa2012,posa2014,chakraborty2014}.
However, to use such models requires tracking their state using real-time perception. 
For rigid-body models and deformable objects such as clothing and towels, there has been a lot of work on tracking the modeled state using sensory feedback~\cite{mordatch2015,schmidt2015,schulman2013}.
For liquids, though, there has not yet been any work connecting physics simulation with real-time perception for robotic tasks.
Unlike modeling rigid or deformable bodies, modeling liquids is much higher dimensional and lacks the same kind of inherent structure, and thus small perturbations can quickly lead to large deviations.
As an example, Figure~\ref{fig:intro} shows a comparison between real liquid (Figure~\ref{fig:ground_truth}) and the result of performing a carefully tuned liquid simulation with the same setup (Figure~\ref{fig:open_loop}).
It is clear that without any feedback, the liquid simulator and the real liquid have significant differences.

%Closing the loop between simulation and reality is important for transferring any type of robotic reasoning from the simulation to the real robot, but it is especially so for liquids.

In this paper, we investigate ways to incorporate sensory feedback into physics-based liquid simulation. By closing the loop between simulation and real-time observations, a robot can  track liquids with much higher accuracy, as illustrated in Figure~\ref{fig:closed_loop}.   Ultimately, the ability to accurately track the state of a liquid will enable a robot to reason about liquids in a wide variety of contexts, addressing questions such as ``How much water is in this container?'', ``Where did this liquid come from?'', ``What is the viscosity of this liquid?'', or ``How can I move a specific amount of this liquid without spilling?''. Toward this goal, our work only assumes that the robot can track 3D mesh models of the objects in its environment and can differentiate between liquid and everything else in its camera observations, both tasks that have been addressed in prior work~\cite{schmidt2014,cifuentes2016,schenckc2016b}.  We demonstrate that our closed-loop liquid simulation enables a robot to reason about liquids in ways that were infeasible before, such as estimating the amount of water in an opaque container during a pouring task, or detecting partial obstruction in water pipes. 

%Overview paper/results
In this paper we first discuss related work, followed by a detailed description of the liquid simulator we use as the base for our closed-loop physics-based model.
Next we describe two different methods for using the observations of real liquid to correct errors in the base liquid simulator.
After that we describe three experiments we performed using this methodology and their results.
We end the paper with a discussion of the implications of our method and future work.

\newlength{\objectsize}
\setlength{\objectsize}{3.5cm}
\begin{figure}[t]
    \centering
    \setlength{\fboxsep}{0pt}
    \setlength{\fboxrule}{1pt}
    \setlength{\unitlength}{1.0cm}
    
    \begin{subfigure}{\objectsize}
        \fbox{\includegraphics[width=\objectsize]{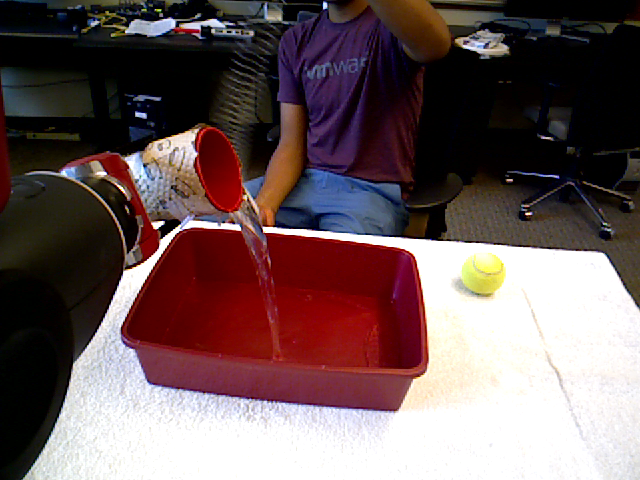}}\vspace{-0.2cm}
        \caption{Color image} 
        \label{fig:color}
    \end{subfigure}\hspace{0.2cm}%
    \begin{subfigure}{\objectsize}
        \fbox{\includegraphics[width=\objectsize]{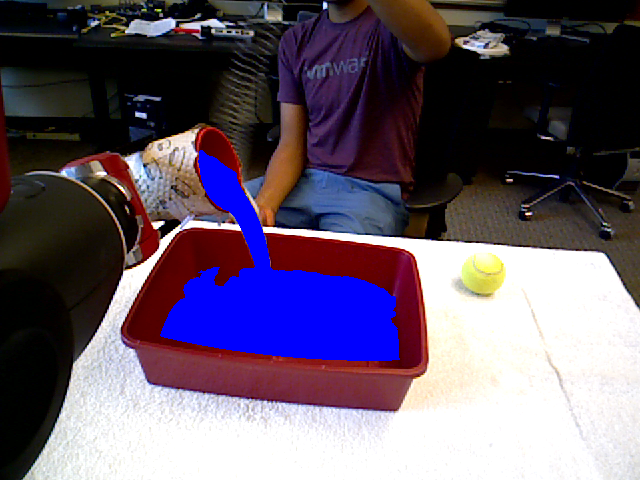}}\vspace{-0.2cm}
        \caption{Ground Truth}
        \label{fig:ground_truth}
    \end{subfigure}
    
    \begin{subfigure}{\objectsize}
        \fbox{\includegraphics[width=\objectsize]{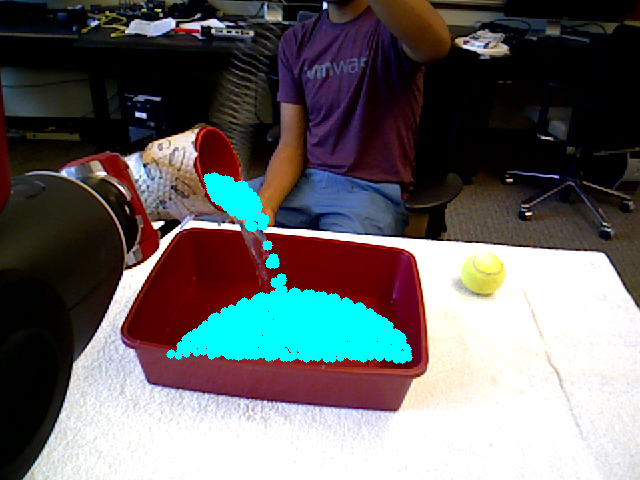}}\vspace{-0.2cm}
        \caption{Open-loop simulation}
        \label{fig:open_loop}
    \end{subfigure}\hspace{0.2cm}%
    \begin{subfigure}{\objectsize}
        \fbox{\includegraphics[width=\objectsize]{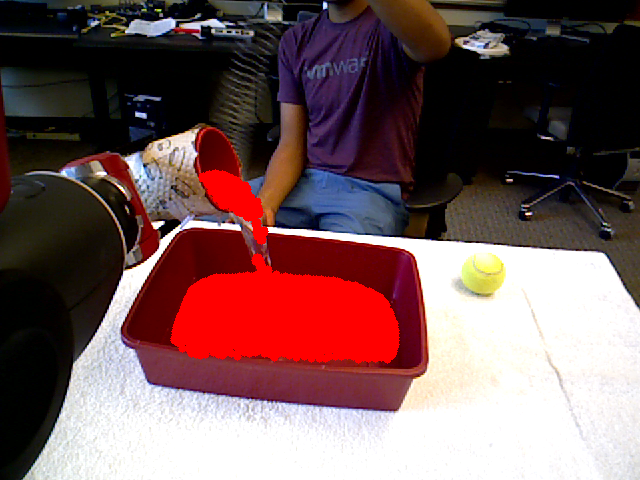}}\vspace{-0.1cm}
        \caption{Closed-loop simulation}
        \label{fig:closed_loop}
    \end{subfigure}

    \caption{ A comparison between open-loop and closed-loop liquid modeling. The upper-left shows the color image of the scene for reference and the upper-right shows the same image with the actual liquid pixels labeled. The lower two images show the color image, but with the liquid from the simulator shown.}
    \label{fig:intro}
    \vspace{-0.8cm}
\end{figure}

\section{Related Work}

Liquid simulation and fluid mechanics are well researched in the literature \cite{acheson1990}.
They are commonly used to model fluid flow in areas such as mechanical and aerospace engineering \cite{hill1992}, and to model liquid surfaces in computer graphics \cite{bridson2015,clavet2005,muller2003}.
Work by Ladick\'{y} {\it et al.} \cite{ladicky2015} combined these methods with regression forests to learn the update rules for particles in a particle-based liquid simulator.
There has also been some work combining real world observations with deformable object simulation.
Schulman {\it et al.} \cite{schulman2013}, by applying forces in the simulator in the direction of the gradient of the error between depth pixels and simulation, were able to track cloth based on real observations.
Our warp field method, described in section \ref{sec:warp_field}, applies a similar concept to liquids.
Finally, the only example in the literature of combining real observations with liquid simulation is work by Wang {\it et al.} \cite{wang2009}, which used stereo cameras and colored water to reconstruct fluid surfaces, and then used fluid mechanics to make the resulting surface meshes more realistic, although they were limited to making realistic appearing liquid flows rather than using them to solve robotic tasks.

In robotics, there has been work using simulators to reason about liquids, although only in constrained settings, e.g., pouring tasks.
Kunze and Beetz \cite{kunze2014,kunze2015} employed a simulator to reason about a robot's actions as it attempted to make pancakes, which involved reasoning about the liquid batter.
Yamaguchi and Atkeson \cite{yamaguchi2015,yamaguchi2016b} used a simulator to reason about pouring different kinds of liquids.
%However, in \cite{kunze2014,kunze2015,yamaguchi2015,yamaguchi2016b} they do not simulate liquids; rather they use small, rigid spheres as a stand-in for liquid.
However, these works use rather crude liquid simulations for prediction tasks that do not require accurate feedback.
Schenck and Fox \cite{schenckc2016b} used a finite element method liquid simulator to train a deep network on the tasks of detecting and tracking liquids. They did not use the simulator to reason about perceived liquid, though.
%In this paper, similar to \cite{schenckc2016b}, we directly simulate liquids.

Yamaguchi and Atkeson  followed up their simulated work with pouring on a real robot \cite{yamaguchi2016c}.
Several others have also performed the pouring task using a real robot \cite{schenckc2016c,langsfeld2014,okada2006,tamosiunaite2011,cakmak2012,rozo2013}.
However, most of these simply dump the entire contents from the source container into the target, bypassing the need to reason in any detail about the liquid dynamics.
Only \cite{schenckc2016c,rozo2013} actually attempted to pour specific amounts of liquid, requiring at least a partial understanding of liquids on the robot's part.

There has been some limited work on perceiving liquids in real data.
Yamaguchi and Atkeson \cite{yamaguchi2016} used optical flow combined with stereo vision to perceive liquid flows in 3D.
Work by Griffith {\it et al.} \cite{griffith2012} used liquids to assist a robot in understanding containers from sensory data.
In both \cite{schenckc2016b,schenckc2016c}, they use deep networks to both perceive liquids in color images and to reason about their behavior.
However, their deep networks are limited to the specific setting they are trained on, and do not have the broad applicability of general liquid simulators.

% Simulation related work
% -SPH vs. FEM
% -Sim paper with decision forests
% -Kunze

% Yamaguchi and pouring stuff papers

% Our prior work (ISER, ICRA)
% Yamaguchi prior work
% Kunze
% Sim paper with decision forests, some other stuff on SPH and FEM
% Pouring papers
% Shane's water stuff

\section{Open-Loop Liquid Simulator}
\label{sec:open_loop}

Our physics-based model is based on a liquid simulator.
The state of the simulator tracks the liquid over time, simulating it forward while observations prevent it from deviating from the real liquid dynamics.
In this section, we describe how the liquid simulator computes the dynamics of the liquid, and in the following section we describe how the observation modifies the liquid state.

To simulate the trajectory of liquid in a scene, the liquid is represented as a set of particles and the Navier-Stokes equations~\cite{acheson1990} are applied to compute the forces on each particle.
The Navier-Stokes equations require certain physical properties of liquid (e.g., pressure, density) to be defined for all points in $\mathbb{R}^3$.
This is implemented using Smoothed Particle Hydrodynamics (SPH)~\cite{violeau2012}, which computes the value of a property at a specific location in space as the weighted average of the neighboring particles.
This is in contrast to finite element liquid simulations~\cite{reddy2014}, which divide the scene into a voxel grid and store the values of the given property at each location in the grid.
One major disadvantage of the finite element simulations is that as the size of the environment grows, the requirements of the voxel grid in both memory and run time grows as $O(n^3)$, making them inefficient for large environments with sparsely located liquids.
This is the case for the simulations in this paper, and so we chose to use SPH, which is better suited to this type of task.
The implementation used in this paper is based off the implementation from the particle simulation library Fluidix~\cite{fluidix2017}.
The rest of this section briefly describes that implementation.

Smoothed Particle Hydrodynamics is essentially a method for representing a continuous vector field of a physical property in space via a discrete set of particles.
It is based around the following equation for evaluating that field at any arbitrary point in space, where $A$ is the physical property in question: 
\[ A(r) = \displaystyle\sum_j m_j\frac{A_j}{\rho_j} W(|r - r_j|, h) \]
where $m_j$ is the mass of particle $j$, $A_j$ is the value stored in particle $j$, $\rho_j$ is the density of particle $j$, $W$ is a kernel function that weights the contribution of each particle by its distance, and $h$ is the cutoff distance for $W$.
In SPH, the mass $m_j$ of each particle is constant, however the density $\rho_j$ is not, and must be computed via the SPH equation above.
That is, the physical value we want to compute $A$ is set to be the density $\rho$, which results in $\rho$ appearing on the right side of the equation twice.
The issue of recurrence (requiring the density to be known in order to compute the density) is handled by the density in the denominator canceling out:
\[ \rho(r) = \displaystyle\sum_j m_j\frac{\rho_j}{\rho_j} W(|r - r_j|, h) = \displaystyle\sum_j m_j W(|r - r_j|, h).\]

To implement a liquid simulation using SPH, each particle must store 6 physical properties: 3D position (without orientation however since particles are infinitesimally small points), velocity, force, mass, density, and pressure.
As stated above, the mass for each particle is constant.
At each timestep, the force is used to update the velocity as follows:
\[ v_i^{t+1} = v_i^t + \frac{f_i^t}{\rho_i}{\Delta}T \]
where ${\Delta}T$ is the amount of simulation time one timestep corresponds to.
The position at each timestep is then updated by the velocity in a similar manner
\[ r_i^{t+1} = r_i^t + v_i^t{\Delta}T. \]
The density of each particle at each timestep is computed using the equation in the previous paragraph.
The pressure is computed as
\[ p_i = c_i^2(\rho_i - \rho_0) \]
where $c_i^2$ is the speed of sound and $\rho_0$ is the reference density of the liquid.

The force is computed by summing the contributions from pressure, viscosity, gravity, and surface tension.
The pressure force at particle $i$ is defined as:
\[ f_i^{pressure} = \displaystyle\sum_j -\frac{m_j}{\rho_j}\left( \frac{p_i}{\rho_i^2} + \frac{p_j}{\rho_j^2} \right) \nabla W(r_i - r_j).\]
The force due to viscosity is
\[ f_i^{viscosity} = \displaystyle\sum_j -\mu\frac{m_j}{\rho_j}\left( \frac{v_i}{\rho_i^2} + \frac{v_j}{\rho_j^2} \right) \nabla^2 W(r_i - r_j) \]
where $\mu$ is the viscosity constant of the liquid (recall that $v_i$ is the velocity of particle $i$).
To compute the surface tension acting on each particle, we must first compute the normal of each particle:
\[ n_i = \displaystyle\sum_j \frac{m_j}{\rho_j} \nabla W(r_i - r_j).\]
Intuitively, the normal $n_i$ for any particle in the center away from the surface of the liquid will have approximately equal contributions from all directions, resulting in the magnitude of $n_i$ being small.
Conversely, for particles near the surface, $n_i$ will have a large contribution from particles in the direction of the interior of the liquid and very little contribution in the direction of the surface, resulting in an $n_i$ with a large magnitude in the direction away from the surface.
The force due to surface tension is computed as
\[ f_i^{tension} = -\sigma\frac{n_i}{|n_i|} \displaystyle\sum_j \frac{m_j}{\rho_j} \nabla^2 W(r_i - r_j) \]
where $\sigma$ is the liquid's tension constant. 
To prevent numerical instability when $|n_i|$ is small, we only compute the tension force when the normal magnitude is greater than a threshold, i.e., the particle is near the surface.

To simulate the flow of liquid in a scene during an interaction, we assume the simulator is given 3D models of the objects that interact with the liquid as well as their 6D poses over the course of the interaction (obtained for example from an object tracking system such as \cite{schmidt2014}).
We initialize the liquid particles in the scene (details on this in section \ref{sec:exp_res}) and simulate the particles forward at each timestep as the simulator tracks the objects' poses.

Our liquid simulator is implemented using the particle simulation  library Fluidix~\cite{fluidix2017}, which efficiently computes particle interactions on the GPU.
We performed a best-first grid search over the space of parameters (e.g., the viscosity constant) to find the set of values that best match the real liquid dynamics.
For each set of parameters in the grid, we used the evaluation criteria described in section \ref{sec:eval} to score them with respect to the data we collected (described in section \ref{sec:data_collection}), and selected the parameters that best fit the real data.
In doing so, we attempted to make our open-loop simulation as close as possible to the real liquid dynamics.
For efficiency reasons, we use between 2,000 and 8,000 particles in our experiments.
For a detailed derivation of Smoothed Particle Hydrodynamics, please refer to~\cite{violeau2012}.

\section{Closed-Loop Liquid Simulators}

While liquid simulators model fluid dynamics based on physical properties, they often don't model every possible force that could affect the liquid; and even the best simulators still have some error relative to real liquids.
Over time, even small errors can lead to a large divergence between real and simulated liquid behavior.
While this may not be a problem in some cases (e.g., in 3D animation it may only be necessary for a liquid to appear realistic), if we wish to use liquid simulation as a robot's internal model of its environment, it must match the real liquid behavior as closely as possible.

One potential method for alleviating this issue is to improve the fidelity of the simulator.
However, this method has many pitfalls.
It requires knowledge of every possible force that could affect the trajectory of the liquid, not only the standard forces such as pressure and viscosity, but also forces for example due to vacuum suction (as in the case of a plunger), which may require modeling additional elements of the environment.
It can also be very brittle, as every property of every object in the environment must be known ahead of time (e.g., the friction constants over the entire surface of every object).
Finally, and most importantly, even if the simulator is almost perfectly accurate, the initial state of the simulator might not be known (e.g., unknown amount of water in a cup), and it will still deviate slightly from reality and thus accumulate drift, which a purely open-loop system has no way to estimate or correct for.

We propose two methods for dealing with noise when tracking real liquid dynamics using a simulator.
Both methods involve closing the loop, that is, utilizing observations of real liquid dynamics in order to better approximate them in the simulation.
The first, inspired by standard Bayes filters in robotics, is a MAP filter, which uses the observation to ``correct'' simulation errors relative to the observation.
The second, based on modeling physical forces in the simulator, applies a warp field that pulls particles toward observed liquid.
We describe these two methods in the following sections.

\subsection{Bridging the Observation and the State}
\label{sec:particle_projection}

\newcommand{\pfunc}[1]{\widehat{#1}}

Before describing our two closed-loop methods, we briefly describe how we map the full 3D state of the liquid simulator into the robot's perception space. In this work, we assume that the robot's camera only provides 2D images labeled with pixel detections, based on the observation that most liquids, especially water, are not detected by depth cameras.
At any timestep $t$, the robot's perception is thus a binary image $I_t$, with pixels labeled as {\it liquid} or {\it not-liquid}.
In order to directly compare the particles representing the 3D liquid state with the 2D image, the pose of  the particles must be projected into the image.
This is done using the following equation:
\[ \pfunc{x}_t^i = Ax_t^i \left(\begin{bmatrix}0 & 0 & 1\end{bmatrix} x_t^i \right)^{-1} \]
where $x_t^i$ is the pose of particle $i$ at time $t$, $\pfunc{x}_t^i$ is that pose projected onto the 2D image plane, and $A$ is the camera intrinsics matrix:
\[ 
\begin{bmatrix}
    FL_x & 0 & PP_x \\
    0 & FL_y & PP_y 
\end{bmatrix}
\]
where $FL$ is the focal length and $PP$ is the principle point of the camera.
When projecting particles into the image plane, we can take into account occlusions by casting a ray from the particle's 3D pose into the camera's 3D pose and checking if it collides with any of the rigid objects in the scene.
Any particle whose ray collides with an object is not included when updating the dynamics of the simulator as there is no way to directly observe that particle.
For the particles that are not occluded, we can compute the distance in 2D space between pixels in the image and liquid particles, which can then be used to inform the dynamics of the liquid simulator.

Additionally, we can use this projection to compute the likelihood of an image, that is, how well the overall set of liquid particles ``explains'' each of the observed pixels.
We define the function $\Phi$ to be the {\it coverage} function that maps a pixel location to the number of particles that cover that pixel.
To compute this, we place a small, fixed radius sphere at each liquid particle location, then project those spheres back into the camera, ignoring occluded spheres.
The value of $\Phi$ at a given pixel location is then simply the number of these spheres that projected onto that pixel.
We use this function in both our closed-loop methods.

\subsection{MAP Filter Simulator}

We use a maximum {\it a posteri} (MAP) filter as one of our closed-loop simulation methods.
We model each particle as its own filter, with its own set of hypotheses, and use the MAP hypothesis at each time step to compute the dynamics.
Let $\mathcal{P}_t$ be a set of liquid particles in a scene at time $t$, $\mathcal{O}_t$ be the objects and their corresponding 6D poses, and $I_t$ be the observation.
We define $\mathcal{S}\left(\mathcal{P}_{t-1}, \mathcal{O}_{t} \right) = \mathcal{P}_t$ to be the function as described in section \ref{sec:open_loop} that computes the state of the liquid particles at timestep $t$ given the previous state of the liquid particles.

At the beginning of each timestep $t$, all the liquid particles are propagated forward in time by one step via $\mathcal{S}$ using the objects and their poses $\mathcal{O}_t$.
Since $S$ is deterministic, we perform the dynamics sampling step in the filter separately. 
Given a liquid particle $x_t^i$, we sample one hypothesis particle $\tilde{x}_t^{i,n}$ for each location in a grid centered at that liquid particle's position.
The grid has dimension $3 \times 3 \times 3$ and the size of each grid cell is set at a small, fixed constant (we use 5mm in this paper).
This results in 27 hypotheses sampled for each liquid particle.

Next we must compute $P(\tilde{x}_t^{i,n} | I_t, \mathcal{P}_t)$, the probability of each hypothesis particle given the observation and the set of liquid particles.  Here, we must condition on all particles in order to take into account that these particles may already ``explain'' a certain liquid pixel.
We first apply Bayes rule
\[ P(\tilde{x}_t^{i,n} | I_t, \mathcal{P}_t) \propto P(I_t | \tilde{x}_t^{i,n}, \mathcal{P}_t) P(\tilde{x}_t^{i,n} | \mathcal{P}_t). \]
For simplicity, we use a uniform prior $P(\tilde{x}_t^{i,n} | \mathcal{P}_t)$ over all hypothesis particles that are feasible, eliminating those that violate physical constraints, such as moving through a 3D object mesh.
Thus, for all feasible hypothesis particles,
\[ P(\tilde{x}_t^{i,n} | I_t, \mathcal{P}_t) \propto P(I_t | \tilde{x}_t^{i,n}, \mathcal{P}_t). \]

When computing $P(I_t | \tilde{x}_t^{i,n}, \mathcal{P}_t)$, what we really want to know, since this is a MAP filter, is which $\tilde{x}_t^{i,n}$ maximizes this probability.
However, the interaction between $I_t$, $\tilde{x}_t^{i,n}$, and $\mathcal{P}_t$ is highly complex and difficult to compute analytically. 
Instead, we approximate this value with an activation function $\Psi$ which
%, which monotonically increases with $P(I_t | \tilde{x}_t^{i,n}, \mathcal{P}_t)$ and thus preserves the argmax over all $\tilde{x}_t^{i,n}$.
we define to be
\[ \Psi(I_t, \tilde{x}_t^{i,n}, \mathcal{P}_t) = \!\!\!\!\! \displaystyle\sum_{j \in liquid(I_t)} \!\!\!\!\! \frac{ W(| \pfunc{\tilde{x}}_t^{i,n} - j_t |, h) }{\Phi(j_t, \mathcal{P}_t) + 1} \]
where $liquid(I_t)$ is the set of all {\it liquid} pixels in $I_t$, $W$ is a kernel function, $\pfunc{\tilde{x}}_t^{i,n}$ is $\tilde{x}_t^{i,n}$ projected onto the image plane (as described in the previous section), $h$ is the limiting radius for $W$, and $\Phi$ returns the coverage of $j_t$ by $\mathcal{P}_t$ (also described in the previous section).
Intuitively, this function sums the number of {\it liquid} pixels around $\tilde{x}_t^{i,n}$, weighted by their distance to $\pfunc{\tilde{x}}_t^{i,n}$ divided by their coverage, i.e., how well explained that pixel is by $\mathcal{P}_t$.
Thus, the more {\it liquid} pixels around a hypothesis particle, the higher its $\Psi$ value, and the less the pixels are covered by the liquid particles, the higher the $\Psi$ value.
For $W$ we use a squared exponential kernel with a length scale of $\frac{1}{33^2}$, and we set the limiting radius to $100$.
Intuitively, this means that the unit length under this kernel is 33 pixels with a limiting radius of 100 pixels.

Finally, we set $x_t^i$ from the MAP hypothesis particles as follows:
\[ x_t^i = \underset{\tilde{x}_t^{i,n}}{\operatorname{argmax}} \; \Psi(I_t, \tilde{x}_t^{i,n}, \mathcal{P}_t). \]
Note that we also adjust the velocity of $x_t^i$ to match the change in position from $x_{t-1}^i$ so as to preserve the correct momentum.

%------------------------------------------------
\subsection{Warp Field Simulator}
\label{sec:warp_field}

The second method we use for closing the loop in the simulator is a warp field, somewhat similar to the approach applied in~\cite{schulman2013}.
Here, the observation applies a force in the simulator that attempts to make the liquid particles better match the observed liquid.
Each observation point is essentially a magnet in the scene, pulling nearby particles towards it.
However, if all observation points pulled with the same amount of force, then particles would tend to clump around a subset of the observation points, leaving other observation points with no nearby particles as the forces from the former cancel out those from the latter.
Thus, the amount of force an observation point applies to nearby particles must vary with the number of nearby particles.
When taken together, all the observation points create a field of forces that warp the particles to better match the real liquid observations.

Once again let $\mathcal{P}_t$ be a set of liquid particles in a scene at time $t$, $\mathcal{O}_t$ be the objects and their corresponding 6D poses, $I_t$ be the observation, and $\mathcal{S}$ be the function that computes the dynamics of the particles for a single timestep.
The force due to the observation warp field is computed as
\[ \pfunc{f}_t^{i,obs} = \!\!\!\!\! \displaystyle\sum_{j \in liquid(I_t)} \!\!\!\!\! \lambda \; \frac{u^{ij}_t}{\Phi(j_t, \mathcal{P}_t) + 1} \; W(|\pfunc{x}_t^i - j_t|, h) \]
where $\lambda$ is the warp constant, $liquid(I_t)$ is the set of all {\it liquid} pixels in $I_t$, $u^{ij}_t$ is a unit vector pointing from particle $\pfunc{x}_t^i$ (projected onto the image plane as described in section \ref{sec:particle_projection}) to liquid pixel $j_t$, $\Phi(j_t, \mathcal{P}_t)$ is the coverage of pixel $j_t$ (described in section \ref{sec:particle_projection}) and $W$ is the same kernel function used in the MAP simulator (with same parameters). The warp constant $\lambda$ adjusts the strength of the warp force, with higher values resulting in a higher warp force and lower values in a lower force.

Again, the coverage of a pixel $\Phi(j_t, \mathcal{P}_t)$ is a measure of how many liquid particles ``cover'' it, that is, how many liquid particles are nearby.
The force applied to each particle by each liquid pixel is divided by that pixel's coverage, thus as more particles cover an observed liquid pixel, it pulls particles to it with less force.
Conversely, pixels that have lower coverage pull particles to them with more force, thus encouraging the simulator to move particles so as to fill the contour of the observed liquid.

The force $\pfunc{f}_t^{i,obs}$ is then projected back into 3D space.
This is done by applying the inverse of the projection described in section \ref{sec:particle_projection}.
Because this is 2D to 3D, the projection has an unspecified degree of freedom.
To compensate for this, we assume that the force vector is in a plane parallel to the image plane in 3D space.
Finally, we apply the SPH equation to smooth the forces across the particles
\[ \bar{f}_t^{i,obs} = \displaystyle\sum_j m_j \frac{f_t^{j,obs}}{\rho_j} W(|r_i - r_j|, h). \]
The resulting force $\bar{f}_t^{i,obs}$ is then added to the other forces described in section~\ref{sec:open_loop} and $S$ is computed as normal.

\section{Experimental Setup}

\subsection{Robot \& Sensors}

The robot used in the experiments in this paper was an upper-torso robot with two 7-DOF arms, each with an electric parallel gripper.
A table was fixed in front of the robot.
To sense its environment, the robot used its Asus Xtion Pro RGBD camera, which recorded both color and depth images at $640 \times 480$ resolution at 30 Hz during each interaction, and its Infrared Cameras Inc. 8640P Thermal Imaging camera, which recorded thermographic images at $640 \times 512$ resolution at 30 Hz during each interaction.
The thermal camera was used in combination with heated water to acquire the ground truth pixel labelings.
The cameras were locked in fixed relative positions and placed just below the robot's head at approximately chest height.

\subsection{Data Collection}
\label{sec:data_collection}

\subsubsection{Pouring}

\setlength{\objectsize}{2.0cm}
\begin{figure}
    \centering
    \setlength{\fboxsep}{0pt}
    \setlength{\fboxrule}{1pt}
    \setlength{\unitlength}{1.0cm}
    
    \begin{subfigure}{\objectsize}
        \fbox{\includegraphics[width=\objectsize]{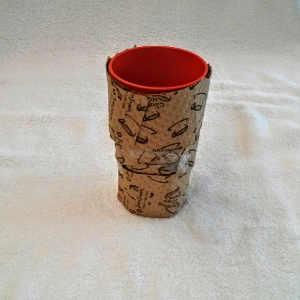}}\vspace{-0.2cm}
        \caption{{\it Cup}}
        \label{fig:cup}
    \end{subfigure}\hspace{0.1cm}%
    \begin{subfigure}{\objectsize}
        \fbox{\includegraphics[width=\objectsize]{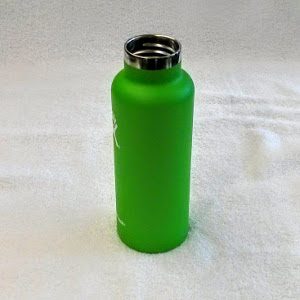}}\vspace{-0.2cm}
        \caption{{\it Bottle}}
        \label{fig:bottle}
    \end{subfigure}\hspace{0.1cm}%
    \begin{subfigure}{\objectsize}
        \fbox{\includegraphics[width=\objectsize]{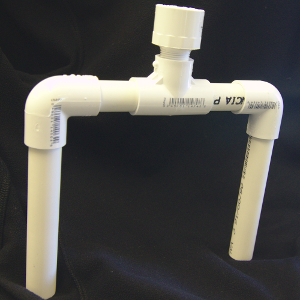}}\vspace{-0.2cm}
        \caption{{\footnotesize\it Pipe Junction}}
        \label{fig:pipe_junction}
    \end{subfigure}
    
    \begin{subfigure}{\objectsize}
        \fbox{\includegraphics[width=\objectsize]{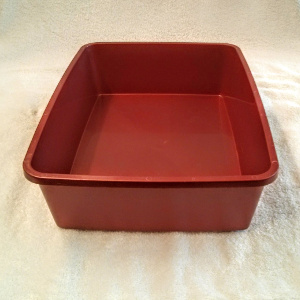}}\vspace{-0.2cm}
        \caption{{\it Pan}}
        \label{fig:pan}
    \end{subfigure}\hspace{0.1cm}%
    \begin{subfigure}{\objectsize}
        \fbox{\includegraphics[width=\objectsize]{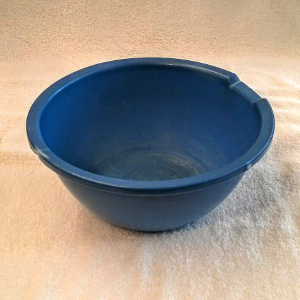}}\vspace{-0.2cm}
        \caption{{\it Bowl}}
        \label{fig:sim_bowl}
    \end{subfigure}\hspace{0.1cm}%
    \begin{subfigure}{\objectsize}
        \fbox{\includegraphics[width=\objectsize]{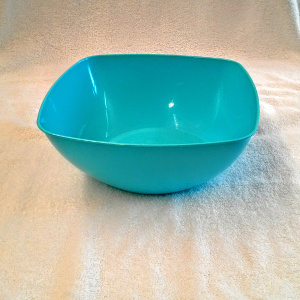}}\vspace{-0.2cm}
        \caption{{\it Fruit Bowl}}
        \label{fig:fruit_bowl}
    \end{subfigure}

    \caption{Objects used during the experiments. The top row shows the two containers the robot poured from as well as the pipe junction. The leftmost bowl in the bottom row was used in the pouring and the right two were used during the pipe junction experiments.}\vspace{-0.2cm}
    \label{fig:objects}
\end{figure}

We collected 16 pouring interactions.
We varied the source container ({\it cup}, Figure~\ref{fig:cup}, or {\it bottle}, Figure~\ref{fig:bottle}) and its initial fill amount (empty, 30\%, 60\%, or 90\% full).
Before each pouring interaction, a bowl (the {\it pan}, Figure~\ref{fig:pan}) was placed on the table in front of the robot.
Next the source was placed in the robot's gripper, filled with water, and the gripper moved over the bowl.
The robot then proceeded to rotate it's wrist along a fixed trajectory such that the opening of the container tilted down towards the bowl and water poured out.
During each pouring interaction, the robot recorded from its RGBD and thermal cameras as well its joint poses.
We collected two trials for each combination of source container and fill amount.

\subsubsection{Pipe Junction}

\setlength{\objectsize}{2.0cm}
\begin{figure}
    \centering
    \setlength{\unitlength}{1.0cm}
    
    \begin{subfigure}{\objectsize}
        \includegraphics[width=\objectsize]{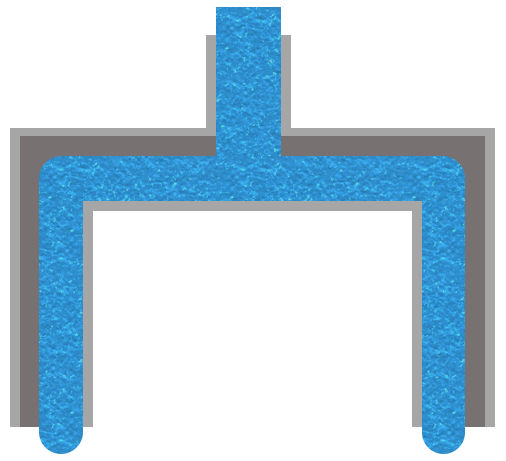}\vspace{-0.2cm}
        \caption{{\it Unblocked}}
        \label{fig:junction_open}
    \end{subfigure}\hspace{0.1cm}%
    \begin{subfigure}{\objectsize}
        \includegraphics[width=\objectsize]{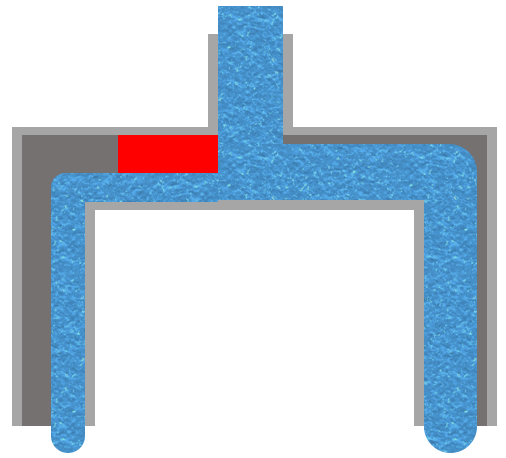}\vspace{-0.2cm}
        \caption{{\it Partial}}
        \label{fig:unction_partial}
    \end{subfigure}\hspace{0.1cm}%
    \begin{subfigure}{\objectsize}
        \includegraphics[width=\objectsize]{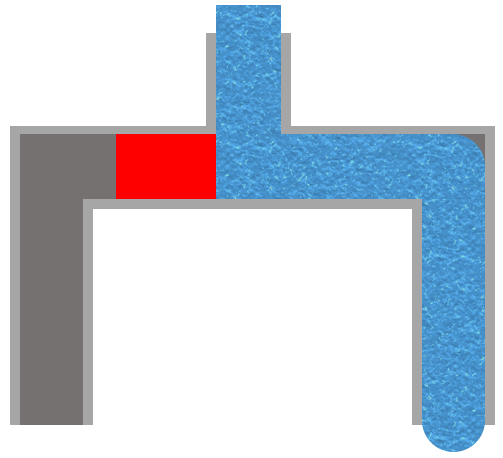}\vspace{-0.2cm}
        \caption{{\it Blocked}}
        \label{fig:junction_closed}
    \end{subfigure}
    
    \caption{The 3 types of blockages placed in the pipe junction. (left to right) Pipe junction with no blockage; left leg is partially blocked; and left leg is fully blocked.}\vspace{-0.3cm}
    \label{fig:pipe_junctions}
\end{figure}

We collected 5 pipe junctions interactions.
Before each of the pipe junction interactions, two bowls ({\it bowl}, Figure~\ref{fig:sim_bowl}, and {\it fruit bowl}, Figure~\ref{fig:fruit_bowl}) were placed side-by-side on the table in front of the robot.
Next, the robot held the ends of the pipe junction (Figure~\ref{fig:pipe_junction}) with its grippers over the bowls and recorded from its RGBD and thermal cameras while 1 liter of water was poured in the top opening.
Each leg of the pipe junction could be fully blocked or partially blocked, i.e., the flow going to that leg could be partially restricted or entirely stopped.
A diagram of the pipe junction and how the blockages affected flow is shown in Figure~\ref{fig:pipe_junctions}.
The blockage can be placed in either leg, for a total of 5 possible configurations.

\subsection{Data Processing}

Before we can use our simulators to track the flow of liquid in the interactions described in the previous section, we must first perform some post-processing on the data.
First, both the open-loop and closed-loop simulators require the object poses to be known over the course of the interaction.
We utilize an object tracking method based on point cloud data to do this.
Second, both closed-loop simulators require an image with pixels labels for the liquid.
We use a thermal camera to acquire this labeling.
In this paper we perform these steps offline, however both are capable of operating in real-time in online situations.

\subsubsection{Object Tracking}

We use the software program DART~\cite{schmidt2014} (Dense Articulated Real-Time Tracking) to track the objects in each interaction.
DART uses depth images to track objects over time.
We initialize the pose of the bowls by using the Point Cloud Library's~\cite{rusu2011} built-in tabletop segmentation algorithm to find the point cluster on the table, and then set their initial pose to the centroid.
We initialize the containers by computing the robot's forward kinematics to find the gripper pose.
Once initialized, DART returns a pose for each object at each point in time over the interaction.

\subsubsection{Liquid Labeling}

\setlength{\objectsize}{2.125cm}
\begin{figure}
    \centering
    \setlength{\fboxsep}{0pt}
    \setlength{\fboxrule}{1pt}
    \setlength{\unitlength}{1.0cm}
    \begin{subfigure}{\objectsize}
        \fbox{\includegraphics[width=\objectsize]{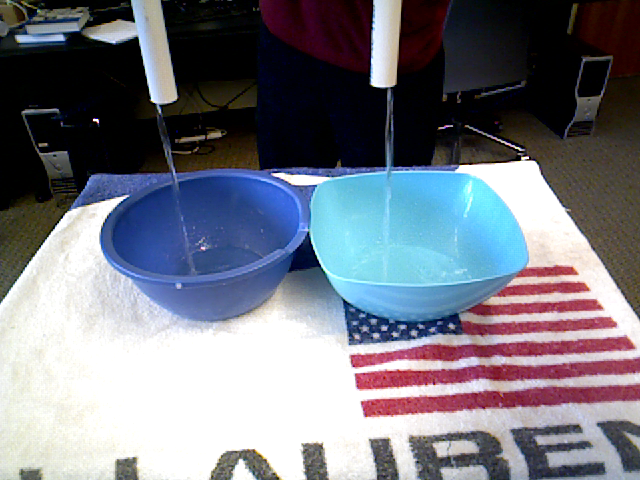}}\vspace{-0.2cm}
        \caption{{\it RGB}}
        \label{fig:calib_rgb}
    \end{subfigure}\hspace{0.1cm}%
    \begin{subfigure}{\objectsize}
        \fbox{\includegraphics[width=\objectsize]{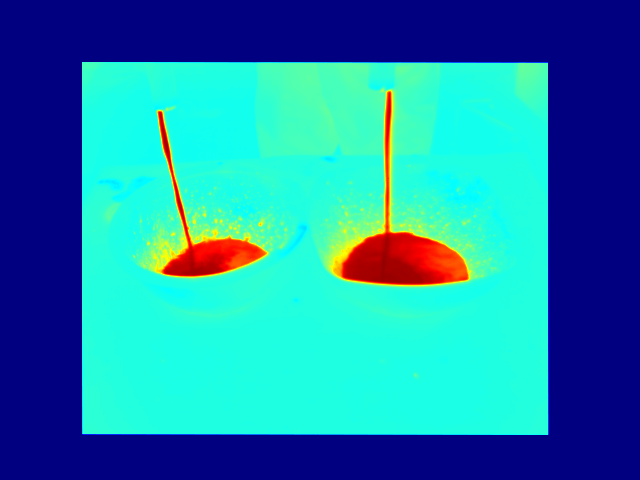}}\vspace{-0.2cm}
        \caption{{\it Thermal}}
        \label{fig:calib_therm}
    \end{subfigure}\hspace{0.1cm}%
    \begin{subfigure}{\objectsize}
        \fbox{\includegraphics[width=\objectsize]{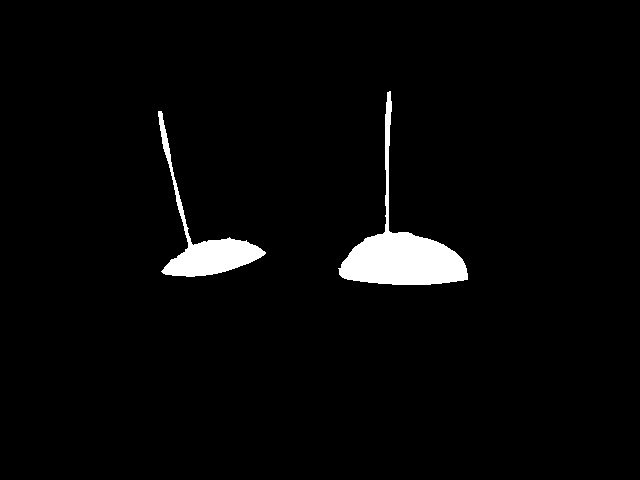}}\vspace{-0.2cm}
        \caption{{\it Threshold}}
        \label{fig:calib_gt}
    \end{subfigure}\hspace{0.1cm}%
    \begin{subfigure}{\objectsize}
        \fbox{\includegraphics[width=\objectsize]{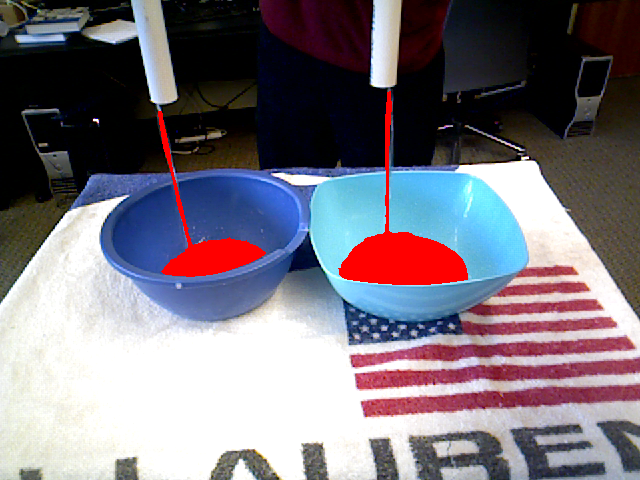}}\vspace{-0.2cm}
        \caption{{\it Overlay}}
        \label{fig:calib_overlay}
    \end{subfigure}
    \caption{Acquiring liquid labels from the thermal camera. The leftmost image is a color image of the scene, the center-left shows the corresponding thermal image transformed to the color image's space. The center-right image shows the liquid labels acquired via thresholding the thermal image, and the rightmost shows the labels overlayed on the color image.}\vspace{-0.3cm}
    \label{fig:thermal_calib}
\end{figure}

For each pouring and pipe junction interaction, the water was heated to a temperature significantly above the surrounding environment but below its boiling point.
The interactions were recorded with a thermal camera, and the thermal image was simply thresholded to locate the liquid pixels.
Figure~\ref{fig:calib_therm} shows an example thermal image recorded during a pipe junction interaction, and Figure~\ref{fig:calib_gt} shows its corresponding thresholded values.

In addition to generating labels from the thermal image, it must also be calibrated to the depth image (the object poses generated by DART, and thus the entire simulator, operate in the depth camera frame of reference).
That is, for each pixel in the thermal camera, we must determine which pixel in the depth camera it corresponds to.
This is not as simple as it may appear.
Water is not visible in the depth image as the projected infrared light does not reflect properly off the surface.
However, our depth camera also collects color images and calibrates it to the depth frame automatically.
We can use the color image then to calibrate the thermal camera.

While there exist methods for doing a full registration between color and thermal images \cite{pinggera2012}, these tend to be noisy and unreliable.
In this paper, because the water remains at a fixed distance from the camera, we use a simpler solution.
First we take a checkerboard pattern printed on a wooden board and place it under a high-intensity halogen lamp.
The light and dark pattern on the board absorbs light from the lamp at different rates, causing the dark squares to heat faster than the light squares.
We then hold this board in front of both the thermal and color cameras at the same distance as the water.
The differential heating causes the checkerboard pattern to be visible in both cameras, allowing us to find correspondence points between the two images.
We then use these points to compute an affine transformation between the images, and use it to transform the thermal image onto the color image.
Figures \ref{fig:calib_rgb} and \ref{fig:calib_therm} show an example color image and its corresponding thermal image transformed onto the color space (the thermal camera has a narrower field of view than the color camera, which is why there are no thermal values around the edge of Figure~\ref{fig:calib_therm}).
Figure~\ref{fig:calib_overlay} shows the thresholded thermal image overlayed onto the color image.

\subsection{Evaluation Criteria}
\label{sec:eval}

We use two criteria for evaluating our methodology.
The first is intersection over union (IOU).
In this case, the state of the liquid simulation is projected into the camera by placing small spheres at each particle location and projecting those into the camera, taking into account occlusions by objects.
We then compare the set of pixels labeled as {\it liquid} by this projection to the set of pixels labeled as {\it liquid} by the thermal image.
The IOU is simply the intersection of these two sets divided by the union.

When comparing the probability of multiple simulations for the purposes of estimating hidden state, we use $P(\hat{\mathcal{I}}_\pi | \mathcal{I}_\pi)$ where $\hat{\mathcal{I}}_\pi$ is a set of predicted images for interaction $\pi$, and $\mathcal{I}_\pi$ is the set of ground truth images.
To compute this, we first apply Bayes rule
\[ P(\hat{\mathcal{I}}_\pi | \mathcal{I}_\pi) \propto P(\mathcal{I}_\pi | \hat{\mathcal{I}}_\pi) P(\hat{\mathcal{I}}_\pi). \]
For our experiments, we assume the prior $P(\hat{\mathcal{I}}_\pi)$ is uniform.
To compute $P(\mathcal{I}_\pi | \hat{\mathcal{I}}_\pi)$, we assume each pixel is independent and simply multiply their individual probabilities together
\[ P(\mathcal{I}_\pi | \hat{\mathcal{I}}_\pi) = \displaystyle\prod_{t=1}^T \displaystyle\prod_j P(j | \hat{j}) \]
where we set $P(j | \hat{j})$ equal to $\delta$ if $j$ and $\hat{j}$ are equal (both {\it liquid} or both {\it not-liquid}), and to $1-\delta$ if they are not.
Due the the large number of pixels across all images and timesteps, we set $\delta = 0.50001$ to prevent underflow\footnote{Even in log-space, values would still periodically underflow with higher values for $\delta$ due to the large quantity of pixels.}.
After computing the probabilities, we then normalize them so they sum to 1.

\section{Experiments \& Results}
\label{sec:exp_res}

We ran three experiments to evaluate our simulators at tracking the state of real-world liquids.
The first utilized the pouring interactions and focused on quantitatively evaluating the open and closed loop simulators.
The second and third experiments test our simulation methods at estimating the state of an unknown variable in the environment.
This is an important ability for a robot, as often liquids are occluded by containers or other objects, forcing robots to reason about the hidden state of the liquids based on outcomes during an interaction, something that is not always necessary during rigid object interactions.
Our second two experiments examine two different cases of hidden state estimation using liquids.

\subsection{Comparing Open and Closed Loop Simulation Methods}

\begin{figure}
\begin{center}
\setlength{\fboxsep}{0pt}
\setlength{\fboxrule}{1pt}
\setlength{\unitlength}{1.0cm}

\begin{subfigure}{4.25cm}
  \scalebox{0.7}{
  \begin{tabular}{r | c | c | c }
   & {\bf Open} & {\bf MAP} & {\bf Warp} \\
   & {\bf Loop} & {\bf Filter} & {\bf Field} \\
   \hline
  {\bf Cup } & 60.17\% & 73.38\% & 75.94\% \\
  {\bf Bottle} & 67.25\% & 77.12\% & 79.41\% \\ \hline
  {\bf 30\%} & 35.56\% & 65.22\% & 67.01\% \\
  {\bf 60\%} & 77.62\% & 79.85\% & 82.80\% \\
  {\bf 90\%} & 77.94\% & 80.69\% & 83.22\% \\
   \hline
   {\bf Overall} & 65.66\% & 76.03\% & 78.41\% \\
  \end{tabular}
  }
\end{subfigure}%
\begin{subfigure}{4.25cm}
  \scalebox{0.5}{
  \begin{tikzpicture}
      \node at (0.0,0.0) {\includegraphics[width=9.5cm]{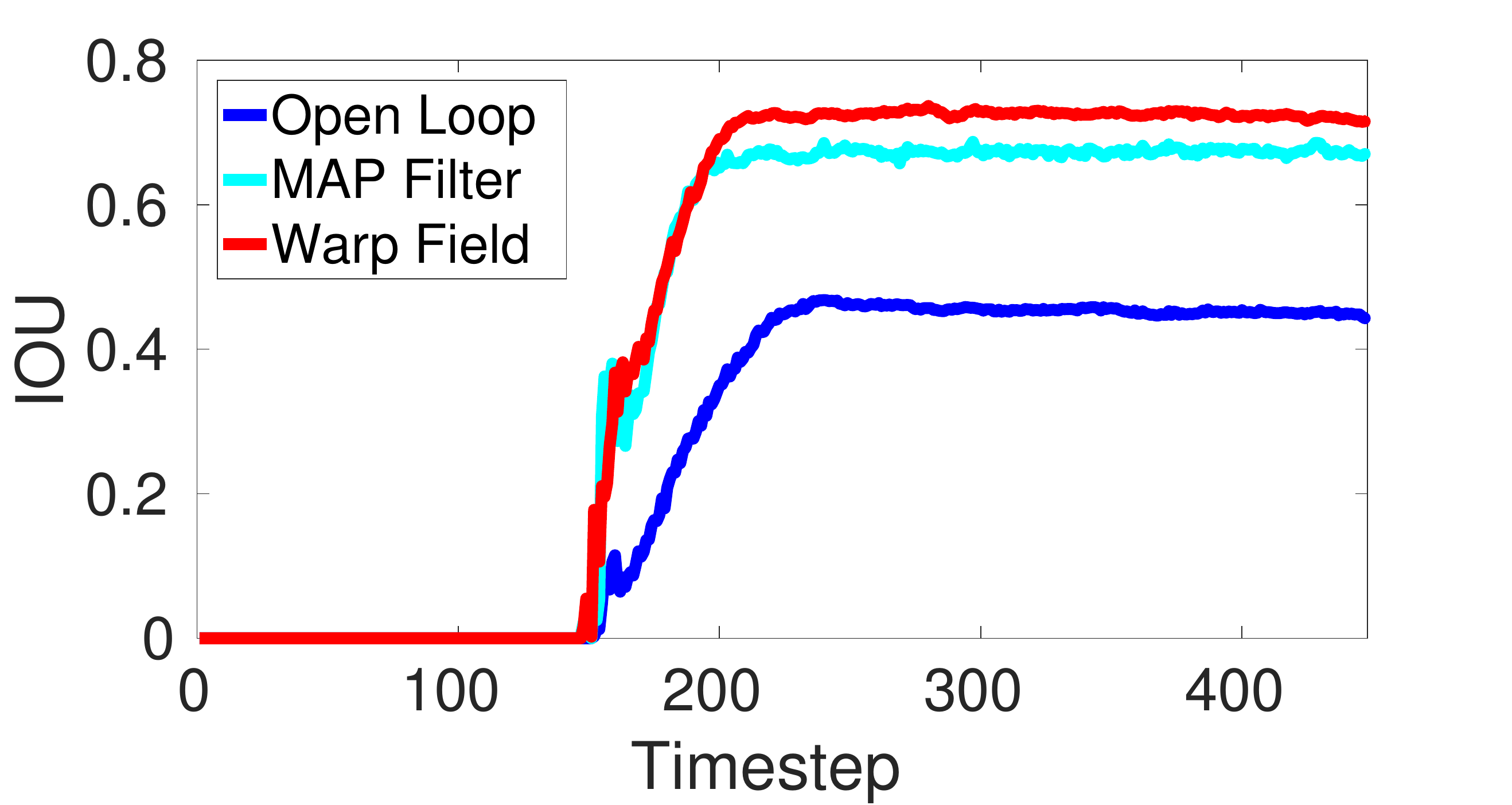}};
      %\node at (0.5,2.4) {\fbox{\includegraphics[width=1.0cm]{FINAL_force_warp.png}}};
      %\node at (0.5,1.1) {\fbox{\includegraphics[width=1.0cm]{FINAL_pf_warp.png}}};
      %\node at (0.5,-0.2) {\fbox{\includegraphics[width=1.0cm]{FINAL_no_warp.png}}};
      %\node at (0.5,-1.3) {\fbox{\includegraphics[width=1.0cm]{FINAL_ground_truth.png}}};
      %\node at (0.5,-0.8) {{\footnotesize\it Ground Truth}};
  \end{tikzpicture}
  }
  %scene_left_pan_bottle_30%_dump_minimal_large_bowls
\end{subfigure}
\caption{The table shows the IOU for each method. The graph shows the IOU at every timestep across one of the pouring experiments ({\it bottle} filled to {\it 30\%}).}
\vspace{-0.5cm}
\label{fig:iou}
\end{center}
\end{figure}

To compare each of the three simulation methods (open loop, MAP filter, and warp field), we simulated them on the data collected for each pouring interaction.
At the start of each interaction, we fill the 3D model of the container with the same amount of liquid as was filled in the real container.
To do this, we perform binary search on the initial number of particles, running the simulation forward, holding the object poses constant, until each has settled and then computing the level of the liquid in the container.
We then simulate the liquid forward in time, updating the object poses based on the tracked poses acquired using DART.
We evaluate each method by comparing their IOUs, computed as described in section \ref{sec:eval}\footnote{The 4 pouring interactions where the container was left empty were not included in this analysis because the union part of the IOU would be 0, resulting in a division by 0.}.

The IOU for the three simulation methods is shown in the table in Figure \ref{fig:iou}.
The upper two rows show the IOU for the methods conditioned on the two types of containers used; The middle rows show the IOU conditioned on the initial percent full of the container; and the last row shows the overall IOU for each method.
This table reveals some interesting phenomena.
It is not immediately clear why all the simulators seem to perform slightly better on interactions where the robot poured from the bottle rather than the cup.
However, the middle of the table shows that all of the methods tend to perform better when more liquid is involved.
We notice that the {\it bottle}, while having a similar diameter as the {\it cup}, is taller, meaning if they are filled to the same ratio full (e.g., 30\%), then the {\it bottle} will have more overall liquid.
This explains the slight bump in performance from one container to the other.

The most important revelation, however, is that both closed-loop simulation methods outperform the open-loop simulation by a significant margin.
This is illustrated graphically by the graph on the right in Figure~\ref{fig:iou}, which shows the IOU at every timestep over one sequence, and clearly shows that the closed-loop methods are better able to match the location of the real liquid than the open-loop method.
Additionally, both the table and the graph show that the warp field method outperforms the MAP filter method.
This clearly shows that closing the loop in liquid simulations can make the trajectory of the liquid better match real world liquid dynamics.

\subsection{Estimating the Initial Amount of Liquid}

\setlength{\objectsize}{4.5cm}
\begin{figure}
    \centering
    \setlength{\unitlength}{1.0cm}
    \begin{subfigure}{\objectsize}
        \includegraphics[width=\objectsize]{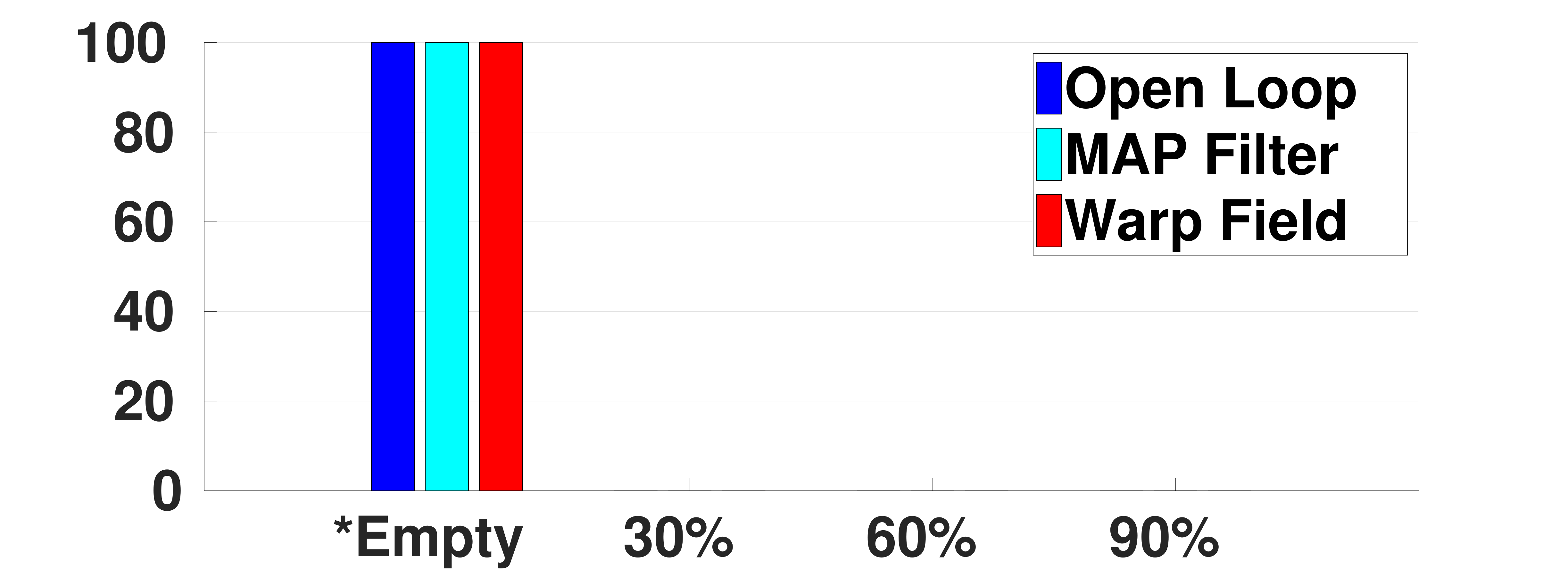}
    \end{subfigure}%
    \begin{subfigure}{\objectsize}
        \includegraphics[width=\objectsize]{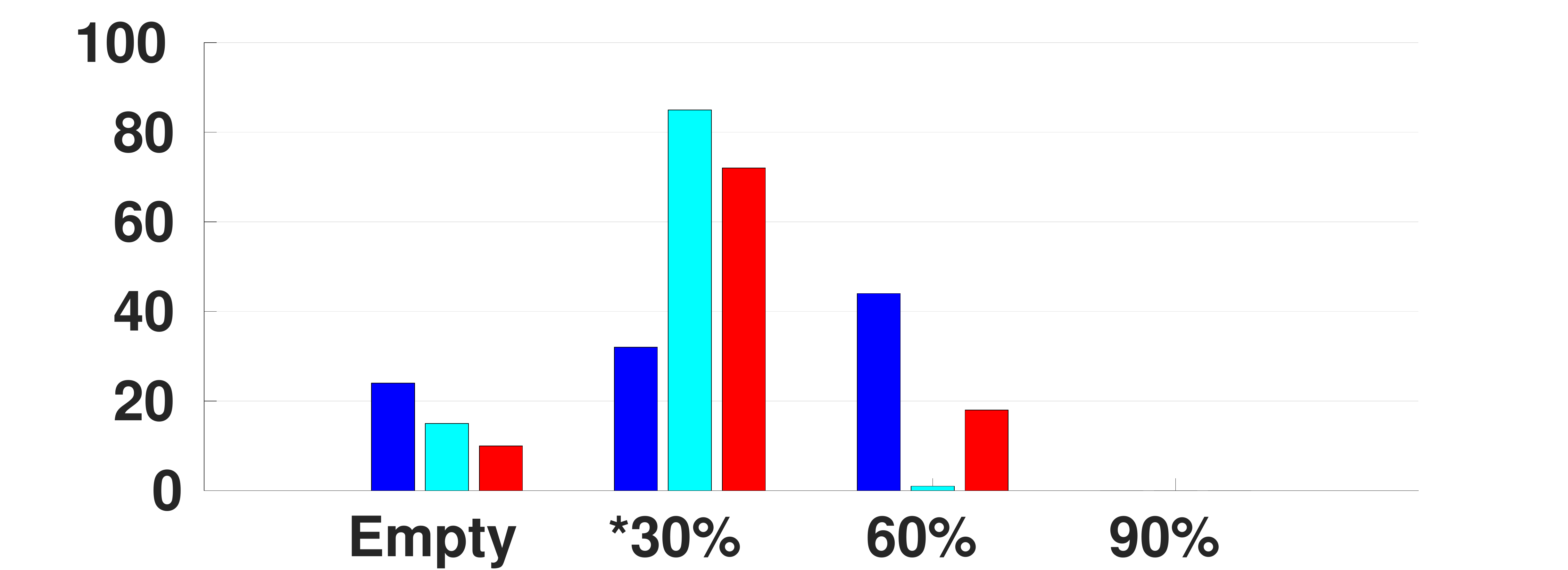}
    \end{subfigure}
    
    \begin{subfigure}{\objectsize}
        \includegraphics[width=\objectsize]{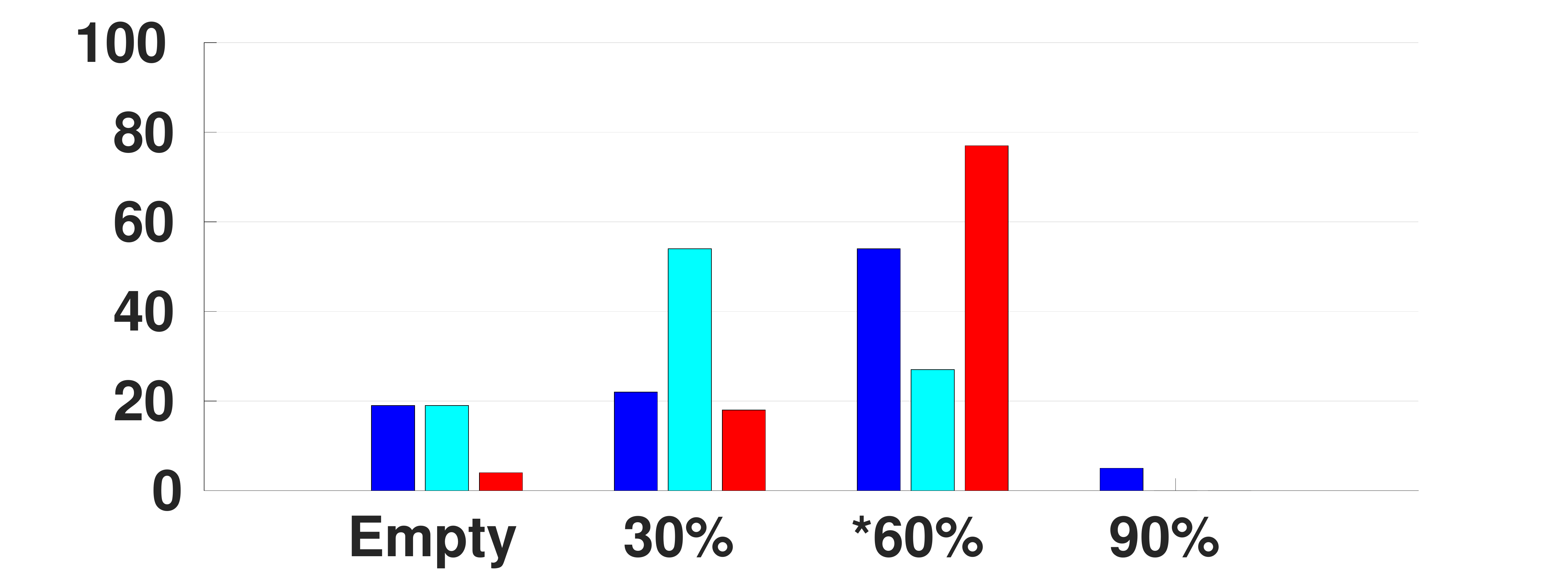}
    \end{subfigure}%
    \begin{subfigure}{\objectsize}
        \includegraphics[width=\objectsize]{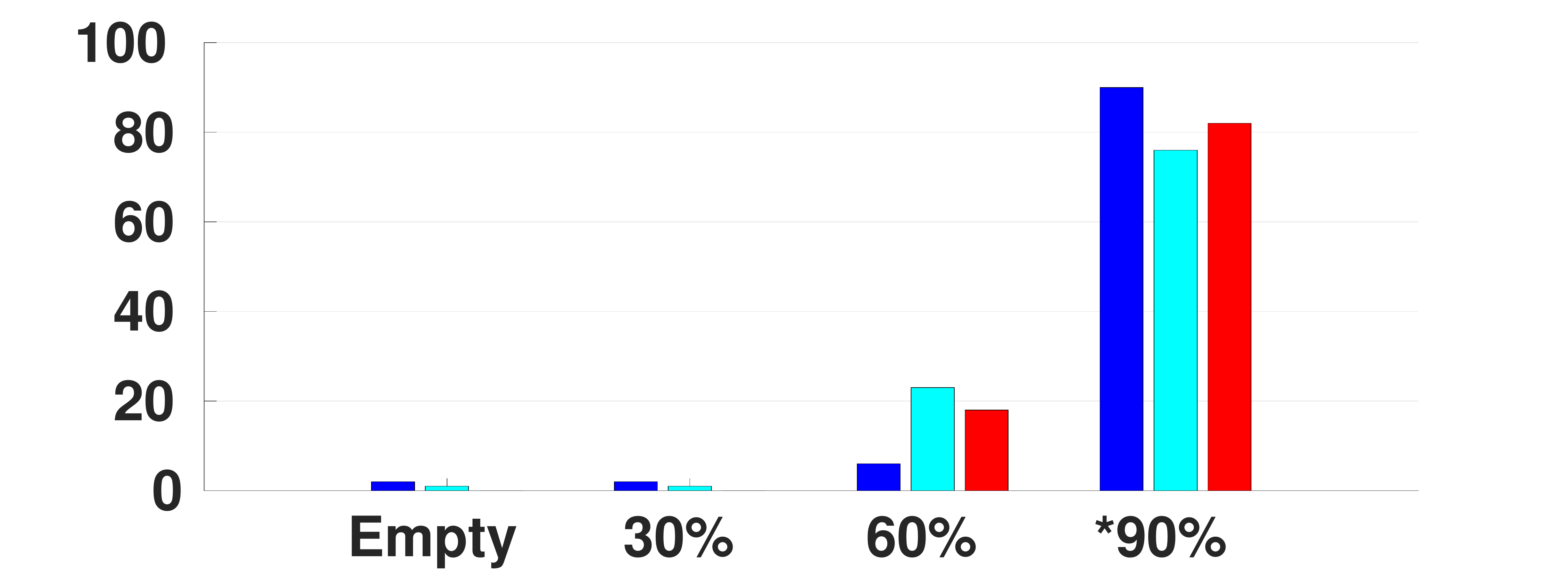}
    \end{subfigure}
    \caption{Probability distribution over the estimated initial fill amounts. They are aggregated by the true fill amounts. From top to bottom they are empty, 30\% full, 60\% full, and 90\% full (indicated by the *). The blue bars show results from the open loop method, cyan for the MAP filter, and red for the warp field.}\vspace{-0.5cm}
    \label{fig:fill_res}
\end{figure}

We evaluated all three simulation methods on the same hidden state task.
For each pouring interaction, the initial amount of liquid in the container was not given to the robot.
Instead, the task of the robot was to estimate this amount based on the observations and its own liquid simulations.
To do this, the robot needs to run multiple simulations for each interaction, one for each possible fill amount, and compare the predictions of each simulation to the observation.

For each pouring interaction, the robot ran 4 simulations: one where the container was left empty, one where the container was filled to 30\% full, one where the container was filled to 60\% full, and one where it was filled to 90\% full.
For each simulation, the liquid particles are simulated forward in time as the object poses are updated via their tracked poses.
We compute the probability of each simulation by evaluating the probability of their predicted images as described in section \ref{sec:eval}.

Figure~\ref{fig:fill_res} shows the results of performing this for each of the pouring interactions, aggregated by the ground truth fill amount (indicated by the * in the x-axis of each graph).
The blue bars show the probability distributions for the open-loop method, the cyan bars show the distribution for the MAP filter method, and the red bars show the distribution for the warp field method.
All methods are easily able to correctly place the highest probability on the empty simulation when there is in fact no liquid in the interaction, which follows intuition as there are no observed liquid particles.
Additionally, even though there is slightly more confusion, all of the methods place the highest probability on the 90\% simulation when the containers start out 90\% full.
Again, this aligns with intuition as it is easy to distinguish ``a lot'' of liquid from ``almost no'' liquid.
The most confusion occurs when trying to distinguish ``a little'' (30\%) from ``some'' (60\%).
The open loop method is almost completely unable to distinguish between the two, both distributions being very similar.
The MAP filter method is slightly better, but still gets confused when the true amount in the container is 60\%.
Only the warp field method is able to correctly estimate the initial amount of liquid, placing over 70\% probability on the correct simulation in every case.

\subsection{Solving the Pipe Junction Task}

The final experiment we performed was the pipe junction task.
Here the task is for the robot to find the blockage in a pair of connected pipes simply by observing the liquid as it exits the pipes, a situation the robot may find itself in if, say, trying to diagnose a broken sink.
We assume that the robot knows {\it a priori} the default, unblocked flow rate of liquid through the pipes, and thus must use the change in flow to find the blockage.
To test this, a pipe T-junction was held inverted over two bowls such that the legs of the T emptied into different bowls, both visible to the robot.
However, the task is to find the blockage based only on the output of the pipes, so the T-junction was held high enough so that the robot could only see the openings on the bottom and not the top opening.
To simulate a constant flow into the pipes, a container with exactly 1 liter of water was tilted at a constant angular velocity so that the liquid flowed into the top opening of the junction.
The type of blockage used (if any), unblocked, partially blocked, or blocked, was placed inside the pipe, not visible to the robot.
We used the data collected during the pipe junction interactions to evaluate the robot on this task.

To solve this task, like in the previous experiment, the robot needs to run multiple simulations with different values for the hidden state (the pipe blockages) and compare their outcomes.
For each interaction, the robot ran 5 simulations: one for both legs unblocked, one for the right leg partially blocked, one for the right leg fully blocked, one for the left leg partially blocked, and one for the left leg fully blocked.
The probability of each simulation is computed using the method described in section \ref{sec:eval}.

Figure~\ref{fig:junction_example} shows the probability for each of the simulated blockages over time for one of the interactions using the best closed-loop method (warp field).
The robot ran one simulation for each blockage type, and the diagrams across the top of the figure indicate where the blockage in that simulation was placed.
The color bordering each diagram corresponds to the color of the line indicating that simulations probability over time.
After only a short time window, the robot is able to place 100\% probability on the correct blockage ({\it partial-left}).
Indeed, we ran this on all 5 pipe junction interactions, and by the end of each, the robot had placed 100\% on the correct blockage in every case.
We also evaluated the 5 interactions using the open-loop method.
It was able to correctly estimate with 100\% probability in the simpler cases (no blockage or fully blocked) as would be expected.
However, for the more difficult interactions (partial blockage), it only picked the correct blockage type and location in one case (when the true blockage was {\it partial-left}) and in the other case incorrectly placed 100\% probability on there being no blockage.
While the point of this experiment was to show the possible type of reasoning that can be done with full physics-based liquid models, even here the closed-loop methods outperform the open-loop methods, if only in 1 out of 5 cases.
Regardless, by using the closed-loop liquid simulation methods developed here, the robot is clearly able to robustly solve this task.

\definecolor{mymagenta}{rgb}{1.0, 0.0, 1.0}
\definecolor{mycyan}{rgb}{0.0, 1.0, 1.0}
\setlength{\objectsize}{1.3cm}
\begin{figure}
    \centering
    \setlength{\fboxsep}{0pt}
    \setlength{\fboxrule}{2pt}
    \setlength{\unitlength}{1.0cm}
    \begin{subfigure}{\objectsize}
        \fcolorbox{blue}{white}{\includegraphics[width=\objectsize]{pipe_junction_open.png}}
    \end{subfigure}\hspace{0.2cm}%
    \begin{subfigure}{\objectsize}
        \fcolorbox{red}{white}{\includegraphics[width=\objectsize]{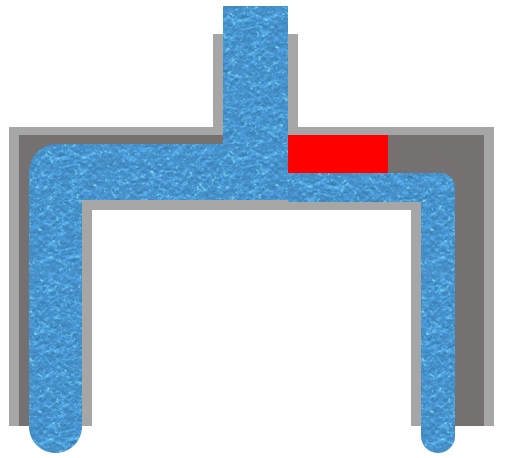}}
    \end{subfigure}\hspace{0.2cm}%
    \begin{subfigure}{\objectsize}
        \fcolorbox{green}{white}{\includegraphics[width=\objectsize]{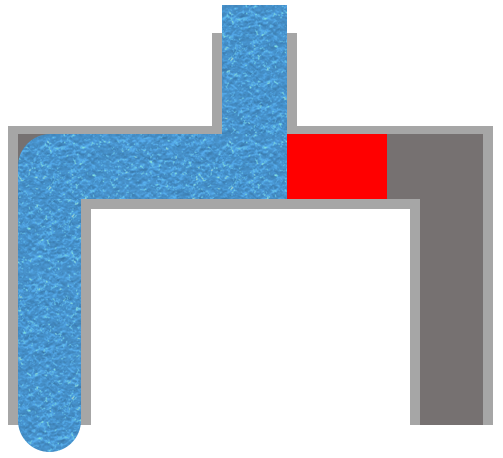}}
    \end{subfigure}\hspace{0.2cm}%
    \begin{subfigure}{\objectsize}
        \fcolorbox{mycyan}{white}{\includegraphics[width=\objectsize]{pipe_junction_partial.png}}
    \end{subfigure}\hspace{0.2cm}%
    \begin{subfigure}{\objectsize}
        \fcolorbox{mymagenta}{white}{\includegraphics[width=\objectsize]{pipe_junction_closed.png}}
    \end{subfigure}
    
    \begin{subfigure}{9.0cm}
        \includegraphics[width=9.0cm]{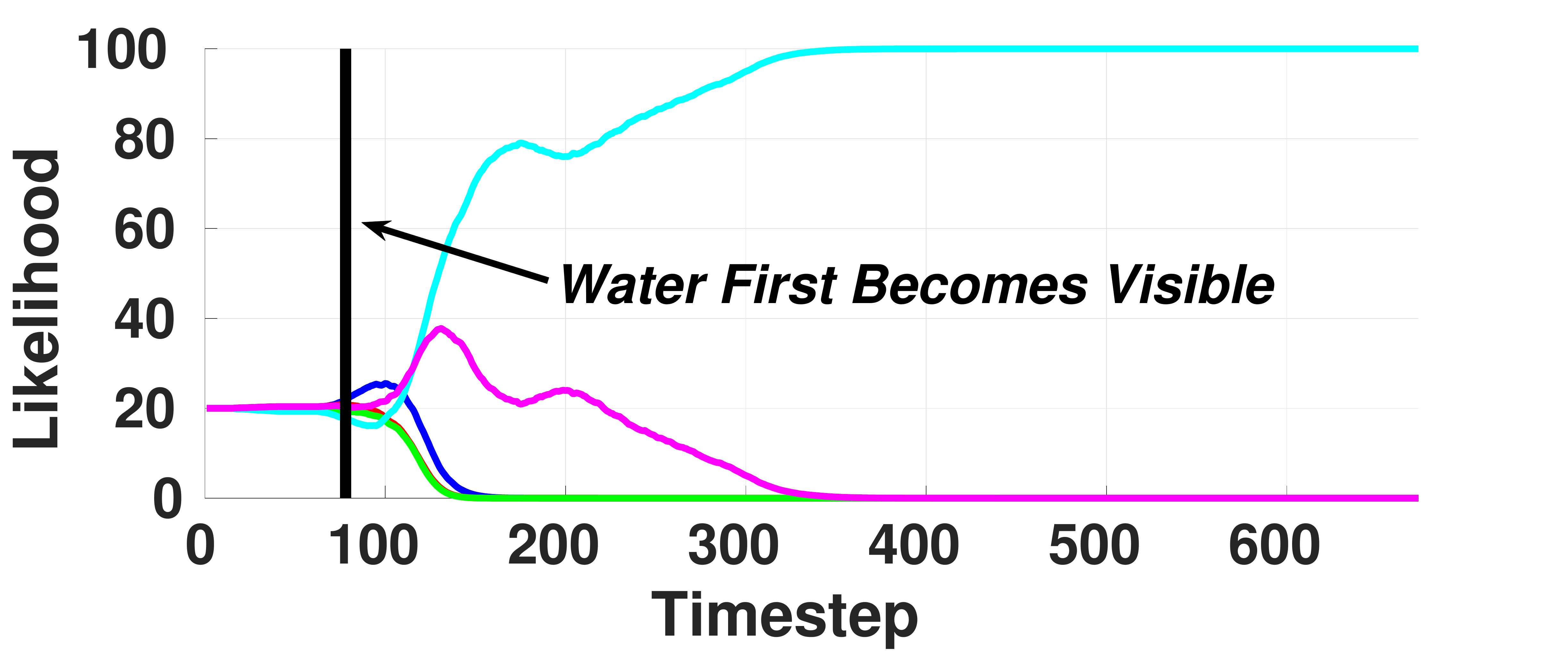}
    \end{subfigure}
    \caption{Probability distribution over the blockage location over time for a single interaction. The 5 diagrams across the top correspond to the five different simulations the robot ran, each color-coded to the corresponding line in the plot. The true blockage was placed in the left leg and only partially blocked the leg (in the keys in the top row, second from right). Best viewed in color.}\vspace{-0.5cm}
    \label{fig:junction_example}
\end{figure}

\section{Discussion}
\vspace{-0.1cm}
% Why is this important? What does it enable?

\textbf{Reasoning about Liquids:} %
So far, reasoning about liquids applied to real robots has been limited to restricted tasks such as pouring \cite{schenckc2016c,rozo2013,cakmak2012}.
With our physics-based model, reasoning about liquids can be done on a much wider variety of tasks.
The last two experiments in this paper both involve completely different tasks, one reasoning about pouring, the other about blockages in pipes, yet the same algorithm is able to solve both tasks, without any special knowledge aside from generic 3D models.
%This is the kind of reasoning that is only enabled through full, physics-based models.
Another advantage of our method over methods such as a deep learning approach \cite{schenckc2016b} or even a non-physics model-based approach \cite{yamaguchi2016} is that the persistence of a liquid is trivially inferred.  
For example, a robot using this model could observe a pouring interaction, and it would be immediately obvious that the new liquid in the target container originated in the source container, and that the overall liquid is the same at the end of the pour as it was at the beginning.

{\bf Generalizing to Other Liquids:} %
Another advantage of a physics-based model is that it can generalize to different types of liquid.
Yamaguchi and Atkeson \cite{yamaguchi2016} developed a model-based detector that could determine the location of liquids in a scene, and they showed that it could generalize to a wide array of liquid types.
This is unlike learning-based models, which cannot generalize to liquids too different from their training set.
With the alteration of a few physical parameters, a physics-based model can generalize to liquids as diverse as water, oil, honey, and even dough.
It is currently an open challenge as to how to infer these parameters efficiently from observation.

{\bf Predicting Liquid Behavior:} %
While others have used physics-based models for liquids \cite{kunze2015}, none have yet combined them with real perception.
As a result, due to the quick divergence of open-loop models with reality, there has been little prior work exploring the possible action spaces around liquids.
Closed-loop liquid simulations enable robots to use the same model to interact with liquids in a wide variety of settings, such as  carrying a container across a room without spilling its contained liquid, scooping liquid with a spoon, and ejecting liquid from a syringe in a controlled manner.
Without closed-loop liquid simulations, each of these tasks would require developing a separate model.
Using an algorithm such as model predictive control \cite{camacho2013}, the robot could plan for a short time horizon into the future using the open-loop simulation, but track the current state using the closed-loop simulation, thus preventing a fatal divergence from reality.

%\vspace{-0.2cm}
\section{Conclusion}
%\vspace{-0.2cm}

In this paper, we proposed two methods for tracking the state of liquid with a closed-loop simulator.
The first, inspired by Bayes filter techniques in robotics, used a MAP filter to correct errors in the simulator.
The second, inspired by the physical forces underlying the simulator, applied a warp field to the particles to correct the error.
The results clearly show that both our closed-loop methods are better at tracking the liquid than the open-loop method.
We also showed how these closed-loop simulations can be used to reason about and infer the hidden variables of an interaction involving liquids.
To our knowledge, this is the first time real liquid observations have been combined with liquid simulations for robotics tasks.

In the immediate future, we plan to continue this work along multiple avenues of investigation.
In this paper, we utilized a thermal camera to acquire liquid detections to focus the evaluation on our experimental methodology.
In the future, we plan to combine our methodology with deep learning methods like the ones in \cite{schenckc2016b,long2015} to perceive liquids, bypassing the need for a thermal camera.
Deep learning can also be applied to perform system identification, i.e., to learn the correct physics models and update them in real-time based on perception.
This might additionally enable more efficient simulation, allowing the use of more particles. Our current system requires running a separate simulator for each hidden state, making it hard to scale to more complex scenarios. One interesting question is how to best incorporate independencies between multiple containers of liquid in order to improve scaling. 
%One of the drawbacks to our method is that, due to efficiency concerns, we only used on the order of thousands of particles, which sometimes resulted in discontinuous liquid in predicted images.
%By using deep learning, it may be possible to increase the efficiency of the simulation, enabling higher resolution liquid modeling.
Additionally, we also plan to apply our methodology to solving closed-loop controls tasks with real liquids, something which was difficult or impossible before.
Finally, we plan to make our data publicly available to other researchers.

%% Use plainnat to work nicely with natbib. 

\bibliographystyle{plainnat}
\bibliography{references}

\end{document}